\pdfoutput=1
\documentclass[10pt,twocolumn,letterpaper]{article}

\usepackage[pagenumbers]{cvpr} 

\usepackage{graphicx}
\usepackage{amsmath}
\usepackage{amssymb}
\usepackage{booktabs}
\usepackage{multirow}
\usepackage{makecell}

\usepackage[pagebackref,breaklinks,colorlinks]{hyperref}

\usepackage[capitalize]{cleveref}
\crefname{section}{Sec.}{Secs.}
\Crefname{section}{Section}{Sections}
\Crefname{table}{Table}{Tables}
\crefname{table}{Tab.}{Tabs.}

\newcommand{\tabincell}[2]{\begin{tabular}{@{}#1@{}}#2\end{tabular}} 

\def\cvprPaperID{} 
\def\confName{CVPR}
\def\confYear{2022}

\begin{document}

\title{PAM: Pose Attention Module for Pose-Invariant Face Recognition}
 
\author{En-Jung Tsai\\
  National Tsing Hua University\\
  {\tt\small enrong.tsai@gapp.nthu.edu.tw}\\
  \and Wei-Chang Yeh\\
  National Tsing Hua University\\
  {\tt\small yeh@ieee.org}\\
}

\maketitle

\begin{abstract}
  Pose variation is one of the key challenges in face recognition. Conventional techniques mainly focus on face frontalization or face augmentation in image space. However, transforming face images in image space is not guaranteed to preserve the lossless identity features of the original image. Moreover, these methods suffer from more computational costs and memory requirements due to the additional models. We argue that it is more desirable to perform feature transformation in hierarchical feature space rather than image space, which can take advantage of different feature levels and benefit from joint learning with representation learning. To this end, we propose a lightweight and easy-to-implement attention block, named Pose Attention Module (PAM), for pose-invariant face recognition. Specifically, PAM performs frontal-profile feature transformation in hierarchical feature space by learning residuals between pose variations with a soft gate mechanism. We validated the effectiveness of PAM block design through extensive ablation studies and verified the performance on several popular benchmarks, including LFW, CFP-FP, AgeDB-30, CPLFW, and CALFW. Experimental results show that our method not only outperforms state-of-the-art methods but also effectively reduces memory requirements by more than 75 times. It is noteworthy that our method is not limited to face recognition with large pose variations. By adjusting the soft gate mechanism of PAM to a specific coefficient, such semantic attention block can easily extend to address other intra-class imbalance problems in face recognition, including large variations in age, illumination, expression, etc.
\end{abstract}

\section{Introduction}
\label{sec:intro}
Face recognition have achieved remarkable progress owing to the development of Deep Learning-based methods.~\cite{taigman2014deepface,sun2014deepid,sun2014deepid2,sun2015deepid2plus,sun2015deepid3}. Previous research have greatly improved the performance of large-scale face recognition by focusing on the design of loss functions~\cite{wen2016discriminative,liu2017sphereface,wang2018cosface,deng2019arcface} for learning discriminative representations. However, several problems still remain in face recognition with unconstrained facial changes, such as head pose, illumination, expressions, and occlusion, which impose great challenges in real-world applications. Among them, head pose has been considered as the most influential intra-class variation on face recognition, as it also leads to the variation of self-occlusions. Sengupta \etal~\cite{sengupta2016frontal} showed that most existing methods have a significant performance gap between frontal-frontal and frontal-profile face verification, while human performance only drops slightly. Recent studies~\cite{zheng2018cross, zheng2017cross} also showed that the existing methods suffer degradation in performance when processing intra-class features with large-pose and large-age variations. This indicates that learning invariant features for intra-class variations remains a great challenge in unconstrained face recognition.

One of the main reasons that large-pose variation confront feature learning challenges is the intra­-class imbalance issues with the training data. The existing datasets have a highly imbalanced proportion between frontal and profile training faces within the same identity. Consequently, deep models tend to learn face representations by more near-frontal training samples rather than evenly distributed frontal-to-profile training samples. This makes modern deep face recognition models unable to properly handle extreme profile faces without further processing. Another key reason is that head rotation induces partial occlusions in crucial facial features. As a two-dimensional image, face images within the same identity may have great feature differences between pose variations. Thus, it is difficult to learn face representation under such feature discrepancies with insufficient and imbalanced training data.

Previous studies have mainly focused on three directions of traditional 3D-based and Generative Adversarial Network-based methods to address the pose variation problem in face recognition, including face frontalization~\cite{hassner2015effective, zhu2015high, huang2017beyond, yin2017towards}, face augmentation~\cite{zhao2017dual, shen2018faceid, deng2018uv}, and pose-invariant representation learning~\cite{tran2017disentangled, zhao2018towards}. These methods synthesize and reconstruct faces in image space, aim to generate photorealistic and identity-preserving images with a canonical view to ease the data imbalanced issues in face recognition. However, transforming face images in image space is not guaranteed to preserve the lossless identity features of the original face image. Moreover, these strategies are based on an additional model that processes the face image before applying it to the recognition model, which significantly increases the burden of the entire system in terms of computational and memory requirements. Therefore, it is necessary to design a more lightweight yet effective solution for pose-invariant face recognition. Lately, Cao \etal~\cite{cao2018pose} proposed a lightweight learning block, named DREAM, that can directly integrate with existing deep face recognition models and learn specific feature transformations in the feature embedding space. Their work is intrinsically similar to face frontalization~\cite{hassner2015effective, zhu2015high, huang2017beyond, yin2017towards} as it performs `frontalization' in the feature embedding space rather than in the image space. This inspires our work to explore more potentialities in hierarchical feature space.

In this study, we hypothesize that all human faces share an approximate feature transformation between pose varia­tions that can be learned from all instances within a large scale dataset. Through this observation, we can mit­igate the problem of intra­-class imbalanced training data by determining this shared approximate feature transformation. Furthermore, in contrast to previous work, we also hypothesize that it is more desirable to perform such feature transformation in hierarchical feature space than in image space and in feature embedding space. By splitting branches from different feature levels of the deep model, it will be more effective to find appropriate features for learning frontal-profile face feature transformation and benefit from joint learning with representation learning.

\begin{figure*}[ht]
  \centering
   \includegraphics[width=0.90\linewidth]{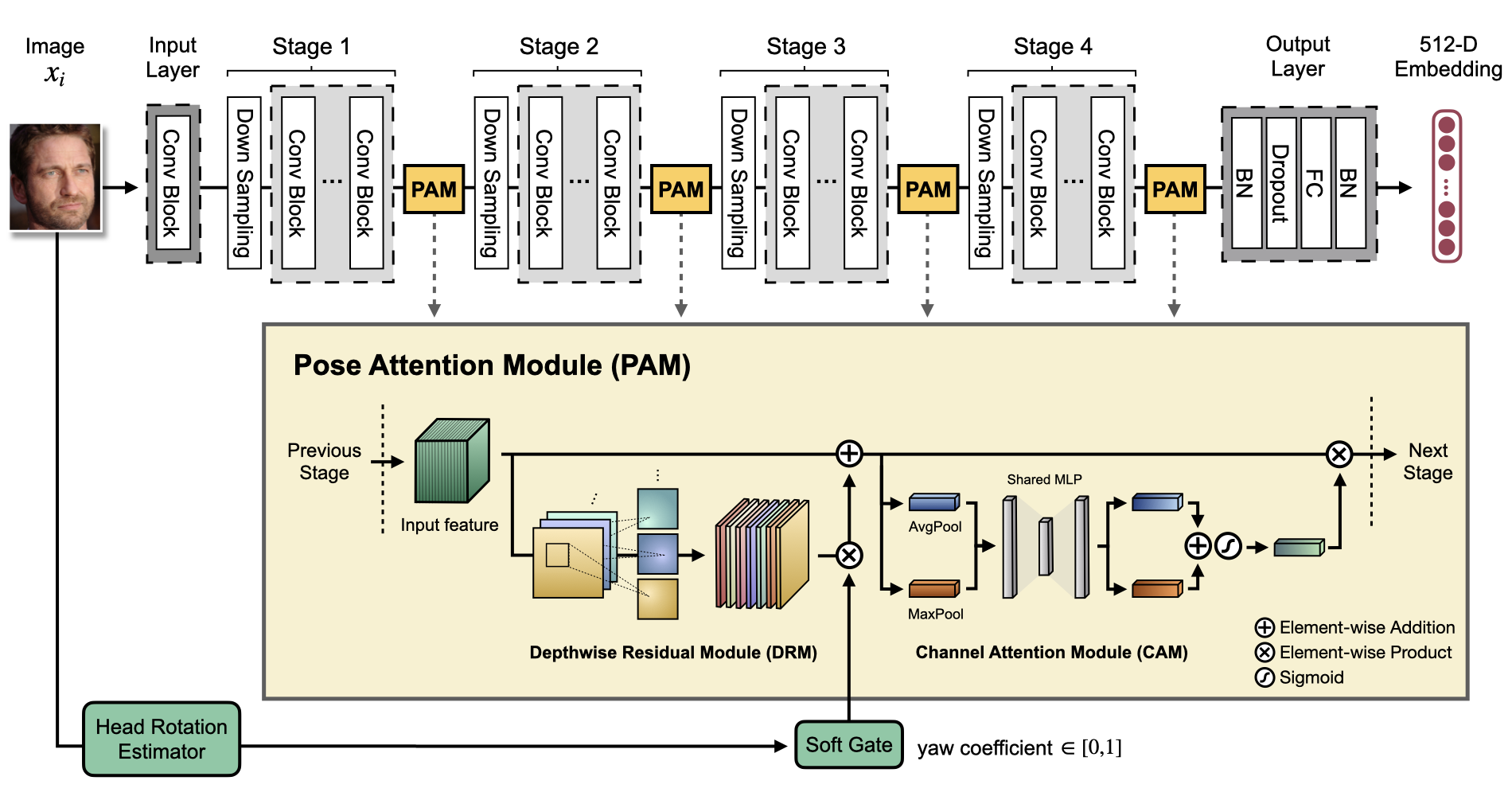}
   \caption{The overall framework of Pose Attention Module (PAM) integrated with the CNNs architectures.}
   \label{fig:overview-of-pam}
\end{figure*}

To this end, we propose a lightweight attention block, named Pose Attention Module (PAM), for pose-invariant face recognition. As shown in~\Cref{fig:overview-of-pam}, PAM performs frontal-profile feature transformation in hierarchical feature space by learning residuals between pose variations with a soft gate mechanism. Specifically, PAM consists of two sequential sub­modules: Depthwise Residual Module (DRM) and Channel Attention Module (CAM). Briefly, DRM learns additional features between pose variations by applying depthwise convolution with a soft gate learning strategy, while CAM improves the representation of interests by reweighting the feature channels after DRM aggregate pose information. The contributions of this work can be summarized as follows:
\begin{itemize}
  \setlength\itemsep{0em}
  \item We propose a novel attention module, named PAM, which can take advantage of different feature levels to better learn feature transformation by interacting with the hierarchical representation during the learning process.
  \item The proposed PAM is extremely lightweight and easy-­to­-implement. The block design effectively reduces memory requirements by more than 75 times and can be integrated and trained with any feed­forward convolutional neural networks in an end­-to-­end manner.
  \item We conduct extensive experiments on various benchmarks. The results demonstrate the effectiveness of PAM on improving pose­-invariant feature learning and the superiority over the state-­of-­the-­art.
\end{itemize}

\section{Related Work}
\label{sec:related-work}
\noindent\textbf{Face Frontalization.} Face frontalization is the process of synthesizing a frontal face image from an arbitrary pose. Early work on face frontalization was mainly based on traditional computer graphics techniques. Most efforts relied on 3D Morphable Model (3DMM)~\cite{blanz1999morphable} fitting methods to project facial appearance between 2D and 3D facial landmarks. Hassner \etal~\cite{hassner2015effective} used a shared 3D surface as a reference model to approximate the shape of all input faces. Zhu \etal~\cite{zhu2015high} proposed a pose adaptive fitting method based on 3DMM and perform high-fidelity face normalization with facial trend inpainting method to alleviate the impact of self-occlusion. However, conventional methods still suffer from severe texture loss and artifacts, which are insufficient for recognition purposes. Recently, Generative Adversarial Networks (GANs)~\cite{goodfellow2014generative} have made advanced progress in face frontalization. Huang \etal~\cite{huang2017beyond} proposed a Two-Pathway Generative Adversarial Network (TP-GAN), which considers both global face structure and local landmark details for photorealistic and identity-preserving frontalization. Yin \etal~\cite{yin2017towards} incorporated 3DMM into GAN and introduced symmetry loss to recover visual occlusion through self-similarity of the left and right halves.

\noindent\textbf{Face Augmentation.}
Recently, with appealing progress in GAN-based face frontalization~\cite{huang2017beyond,yin2017towards}, some works extend face synthesizing to the data augmentation task that generates faces with arbitrary poses. Zhao \etal~\cite{zhao2017dual} proposed DA-GAN to refine the synthesized profile images, enabling conventional 3DMM-based methods to balance the pose distribution of the dataset with more photorealistic and identity-preserving images. Shen \etal~\cite{shen2018faceid} proposed a three-player GAN with a criterion of information symmetry to ensure the real and synthesized images are projected into the same feature space. Deng \etal~\cite{deng2018uv} introduced an adversarial UV completion method, named UV-GAN, which can learn an identity-preserving complete facial UV map from in-the-wild images. By attaching the completed UV map to the fitted mesh, an instance of arbitrary poses can be generated to increase pose variations.

\noindent\textbf{Pose Invariant Representation Learning.} Since GAN-based methods have proven their effectiveness in synthesizing frontal face images~\cite{huang2017beyond,yin2017towards}, some works tend to consider both frontalization and recognition tasks jointly to benefit from each other. Tran \etal~\cite{tran2017disentangled} considered a unified network that jointly learns face frontalization and representation learning together and proposed DR-GAN~\cite{tran2017disentangled}, which extends GAN with a disentangled encoder-decoder structure that can learn pose-invariant representation through pose code conditioning. By taking advantage of prior works, Zhao \etal~\cite{zhao2018towards} proposed a PIM framework that inherits the joint learning process of DR-GAN~\cite{tran2017disentangled} and the dual-path structure of TP-GAN~\cite{huang2017beyond}. The proposed PIM~\cite{zhao2018towards} applies a novel cross-domain adversarial training and siamese discriminator with dynamic convolution sharing weights, which effectively enhance the generalizability of pose-invariant face recognition.

Although the aforementioned methods provide a promising way via generation, however, transforming face images in image space is not guaranteed to preserve the lossless identity features of the original face image. Most methods suffer either unreal or feature loss under large poses due to self­-occlusion, which indicates that de­tails of the transformed face are only practical for better visualization. Moreover, these strategies tend to suffer more computational costs and memory require­ments due to the additional models. 

Lately, Cao \etal~\cite{cao2018pose} present a representative work, named Deep Residual EquivAriant Mapping (DREAM), which is a lightweight learning block that can directly integrate with ex­isting deep face recognition models. DREAM exploits a soft gate learning strategy to model an additional transformation between frontal-profile faces in feature embedding space. Their work is intrinsically similar to face frontalization~\cite{hassner2015effective, zhu2015high, huang2017beyond, yin2017towards} as it performs `frontalization' in the feature embedding space rather than in the image space. Although their results are encouraging, we argue that it is more effective and desirable to perform such feature transformation in hierarchical feature space than in feature embedding space, as it can take advantage of different feature levels. Furthermore, with deliberate block design, it can also inherit the advantage of the shared­-weight structure of convolutional kernels, since the pose variation between frontal-profile faces is also an image task. This inspires our work to explore more potentialities in hierarchical feature space.

\section{Pose Attention Module}
\label{sec:proposed-method}
In this section, we describe the proposed Pose Attention Module (PAM) for pose-invariant face recognition. The block design of the proposed PAM consists of two sequential sub-modules: Depthwise Residual Module (DRM) and Channel Attention Module (CAM). Briefly, DRM learns additional features between pose variations by applying depthwise convolution with a soft gate learning strategy, while CAM improves the representation of interests by reweighting the feature channels after DRM aggregates pose information. ~\Cref{fig:overview-of-pam} illustrates the overall framework of the proposed Pose Attention Module (PAM) integrated with the CNNs architecture. The framework constructs pose attention in hierarchical feature space by placing multiple PAMs at different bottlenecks. We now present each component in detail.

\begin{figure}
  \centering
  \captionsetup[sub]{font=scriptsize}
  \begin{subfigure}{0.483\linewidth}
    \includegraphics[width=\linewidth]{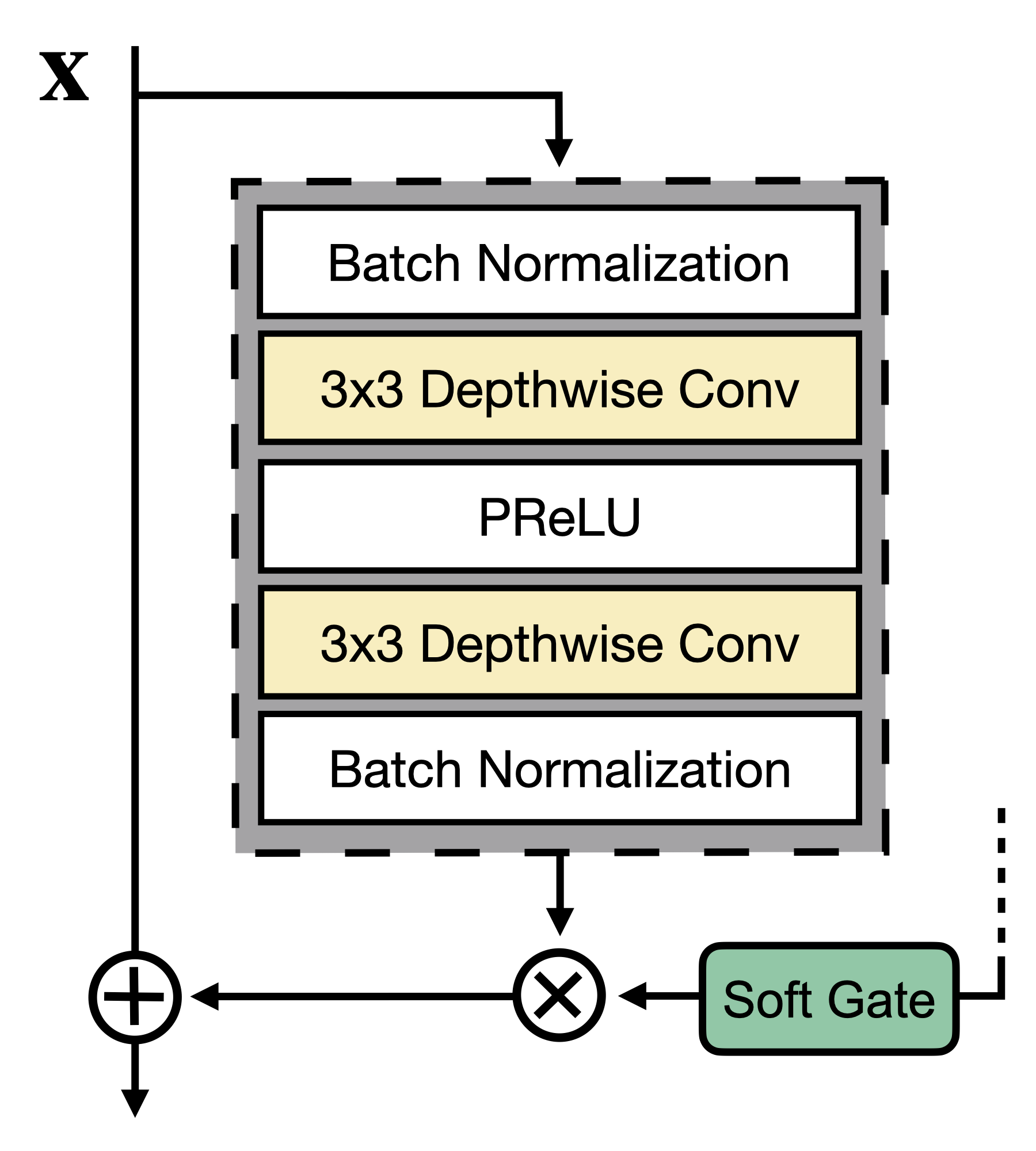}
    \caption{Depthwise Residual Module (DRM)}
    \label{fig:depthwise-residual-module}
  \end{subfigure}
  \hfill
  \begin{subfigure}{0.483\linewidth}
    \includegraphics[width=\linewidth]{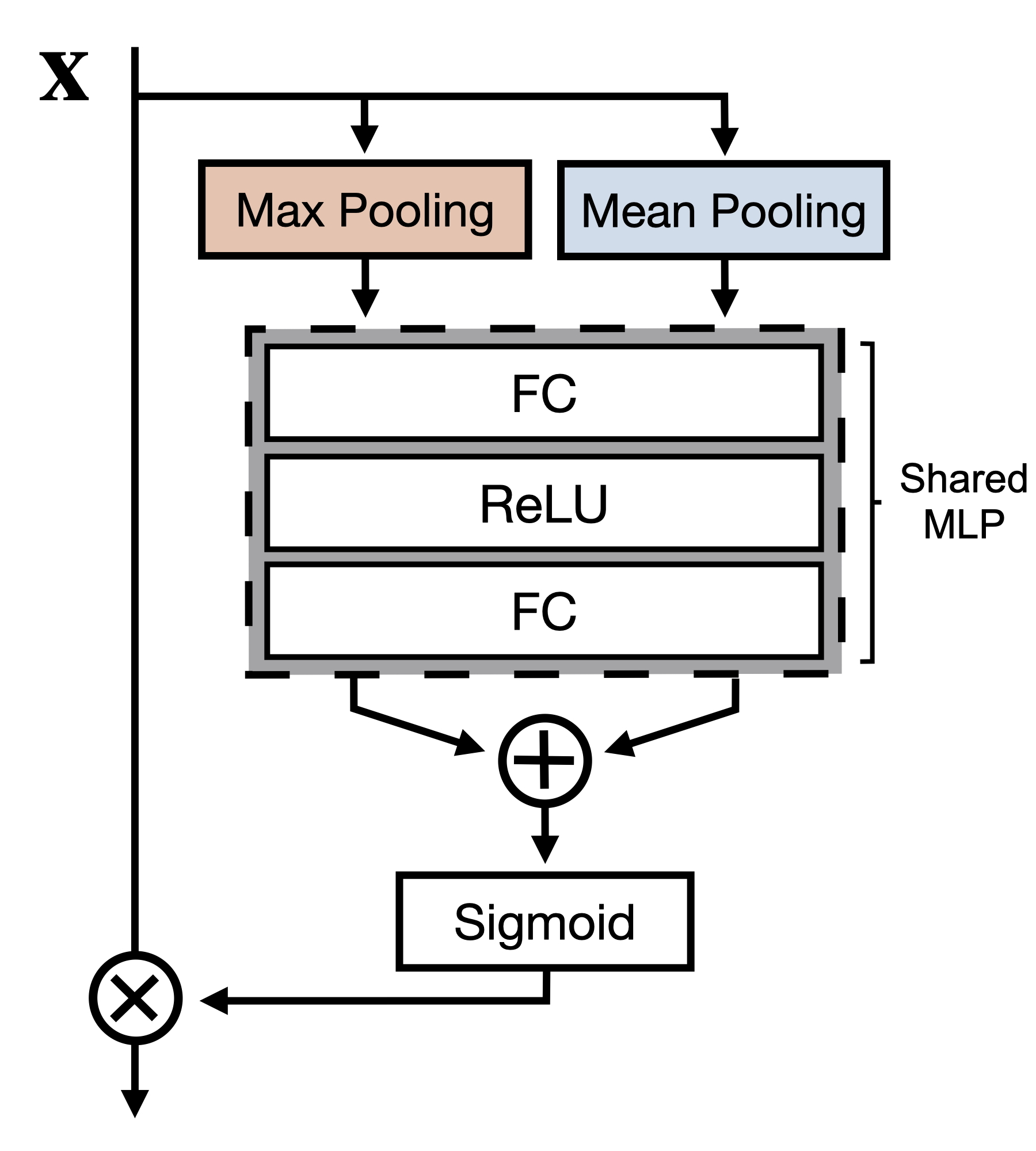}
    \caption{Channel Attention Module (CAM)}
    \label{fig:channel-attention-module}
  \end{subfigure}
  \caption{The block design of the PAM.}
\end{figure}

\subsection{Depthwise Residual Module}
\label{sec:DRM}
The Depthwise Residual Module (DRM) consists of two components: a residual block that learns the residuals between frontal-profile features and a soft gate mechanism that controls the magnitude of the residuals through the yaw coefficient, as shown in~\Cref{fig:depthwise-residual-module}. Specifically, the residual block of DRM has a BN-DepthConv-PReLu-DepthConv-BN structure. In contrast to most residual units~\cite{he2016deep,han2017deep}, we have deliberately replaced the regular convolutional layer with a depthwise convolutional layer for two reasons: feature independence and module efficiency.

\begin{figure}[t]
    \footnotesize
    \begin{minipage}{0.44\linewidth}
        \centering
        \includegraphics*[width=\linewidth]{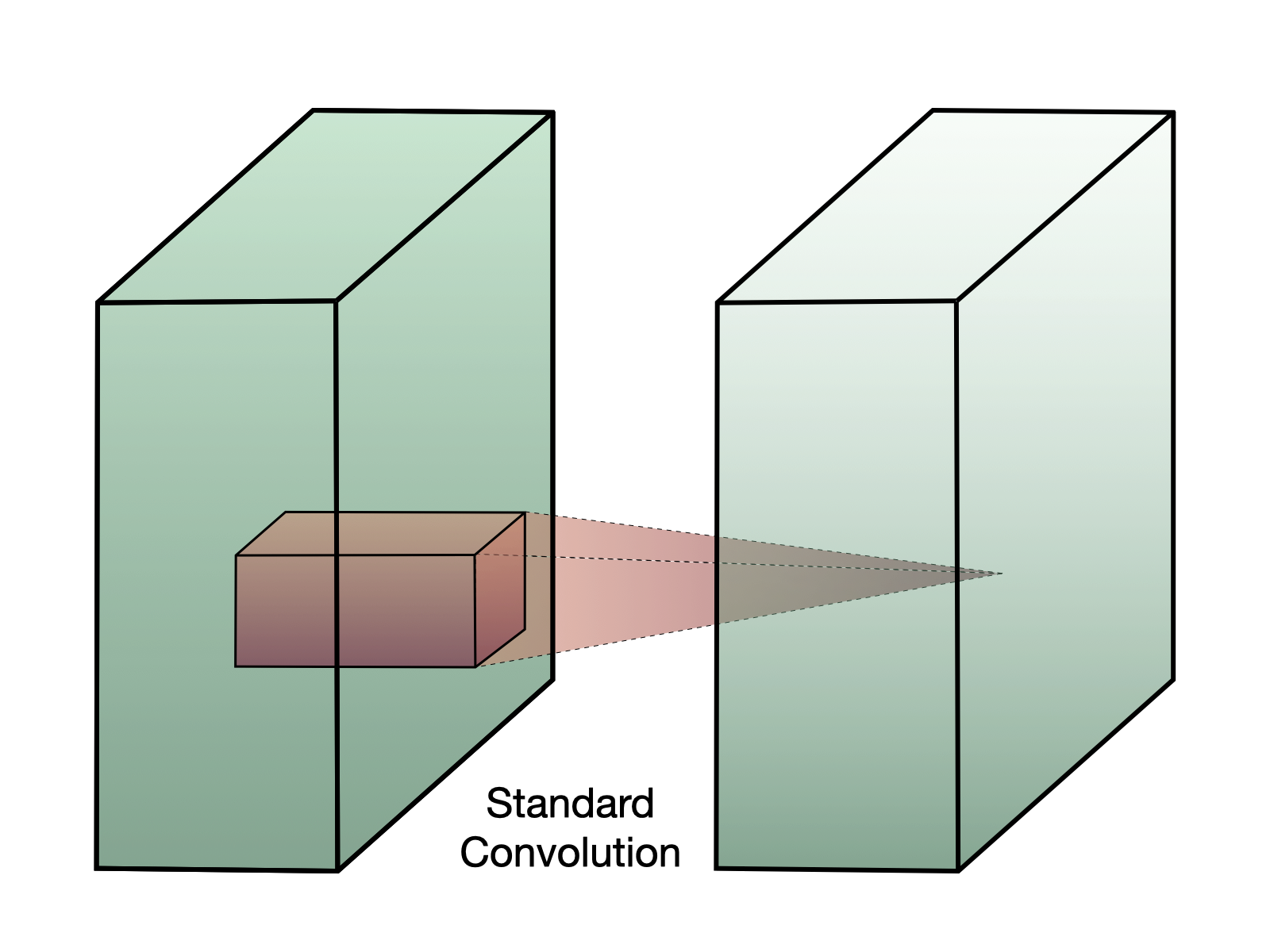}
        (a) Regular Convolution
    \end{minipage}
    \begin{minipage}{0.555\linewidth}
        \centering
        \includegraphics*[width=\linewidth]{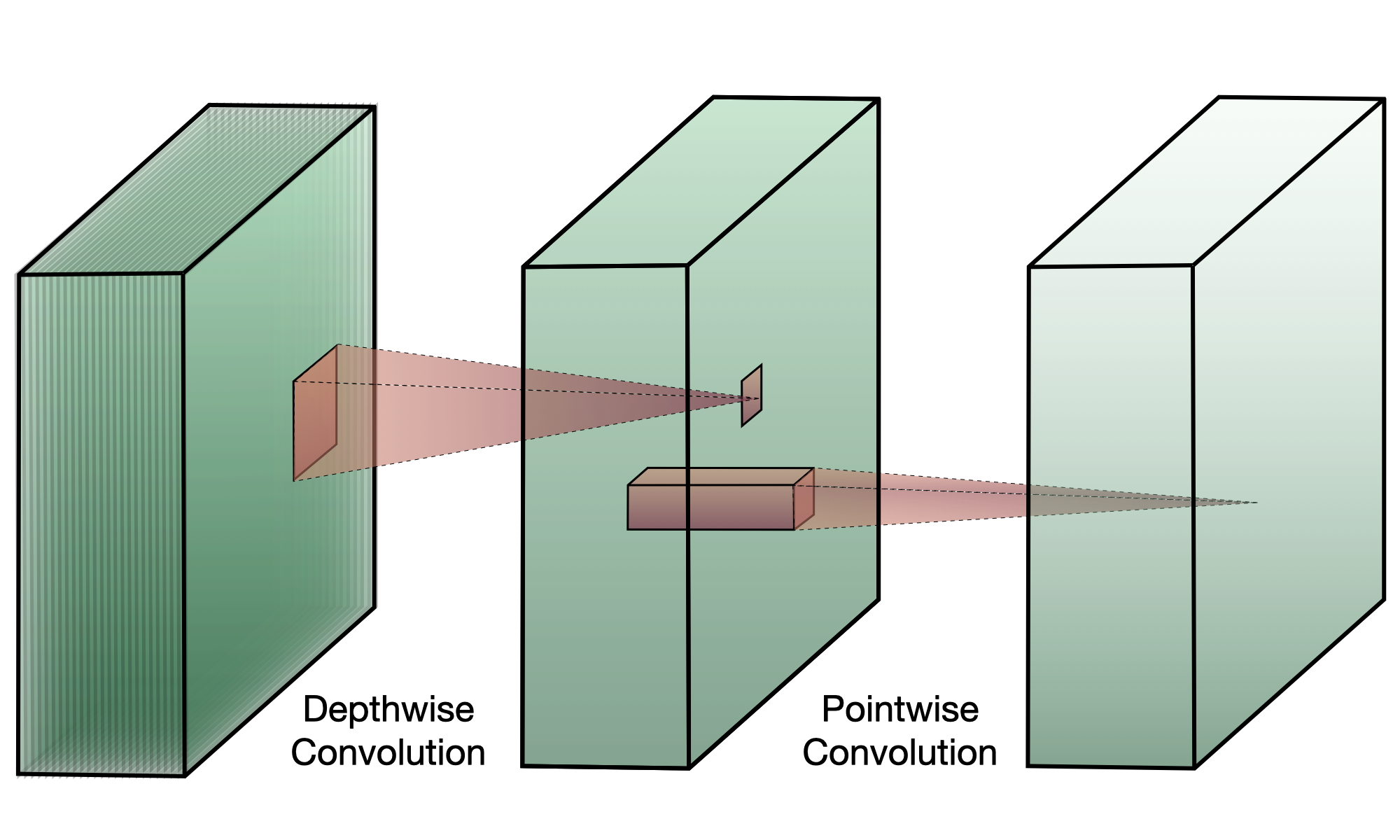}
        (b) Depthwise Separable Convolution
    \end{minipage}
    \normalsize
    \caption{Illustration of regular convolution and depthwise separable convolution~\cite{sandler2018mobilenetv2}.}
    \label{fig:comparison-of-convolution}
\end{figure}

\noindent\textbf{Feature Independence.} In contrast to general residual learning, we designed DRM as a feature transformation of input feature maps. We argue that the abstract feature learned in each channel have different correlations with pose variations, which indicates that DRM should learn feature transformation separately for each channel. Therefore, we applied depthwise convolution to obtain different frontal-profile feature transformations for each channel. As shown in~\Cref{fig:comparison-of-convolution}, depthwise convolution is one of the operations in depthwise separable convolution, which is a common building block for many efficient neural network architectures~\cite{chollet2017xception,howard2017mobilenets,zhang2018shufflenet,sandler2018mobilenetv2}. We verified the performance of different convolution settings by ablation studies (see~\Cref{sec:ablation-study} for more details), where depthwise convolution works slightly better than standard convolution in our experiments.

\noindent\textbf{Module Efficiency.} Given an input tensor $\mathbf{T}_{\mathbf{i}} \in \mathbb{R}^{H_{i} \times W_{i} \times C_{i}}$, a standard convolution produce an output tensor $\mathbf{T}_{\mathbf{j}} \in \mathbb{R}^{H_{i} \times W_{i} \times C_{j}}$ with kernels $\mathbf{K} \in \mathbb{R}^{k \times k \times C_{i} \times C_{j}}$, where $H$, $W$, and $C$ are the height, width and number of channels of $\mathbf{T}_{\mathbf{i}}$ and $\mathbf{T}_{\mathbf{j}}$, and $k$ denotes the kernel size. The computational cost of the standard convolution is $H_{i} \times W_{i} \times C_{i} \times C_{j} \times k^2$ multiplications. Compared to the regular convolutional layer, depthwise convolution can significantly reduce computational cost and memory requirements. The overall consumption of depthwise convolution only cost $H_{i} \times W_{i} \times C_{i} \times k^2$ multiplications. The ratio of computation costs between depthwise convolution and the standard convolution is $\frac{1}{C_{j}}$, where $C_{j}=\{64, 128, 256, 512\}$ for each stage of the recognition model, respectively. 

In conclusion, our method proposed a Depthwise Residual Module (DRM) with a lightweight and effective nature, summarized as follows:
\begin{itemize}
    \setlength\itemsep{0em}
    \item We deliberately apply depthwise convolution in DRM, significantly reducing the computational cost by 64 to 512 times compared to conventional convolution operations.
    \item Extensive experiments show that depthwise convolution brings more improvement than regular convolution in frontal-profile feature transformation, which indicate that learning abstract features independently for each channel is more robust for learning feature transformations.
\end{itemize}

\subsection{Soft Gate Mechanism}
\label{sec:softgate}
The soft gate mechanism plays a crucial role in emphasizing informative features with extreme pose variation and suppressing frontal view signals, which can be considered as an attention mechanism that takes the yaw coefficient as information aggregated from image space. Through this supervision, the DRM tended to learn more profile features as the yaw coefficient increased. By adding the learned residuals to perform frontal-profile feature transformation, the proposed DRM can eventually convert the face features of an arbitrary pose into the pose-invariant features.

\begin{figure}[t]
\centering
\includegraphics[width=\linewidth]{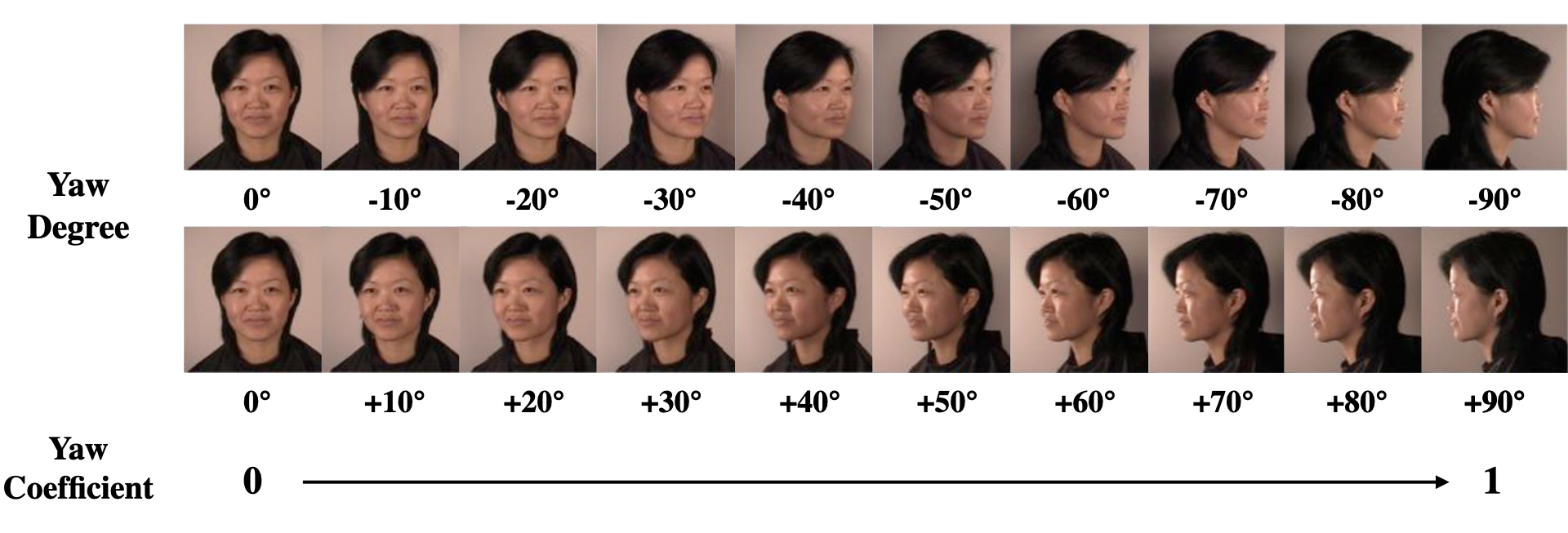}
\caption{Illustration of the yaw coefficient without considering left-right profile direction using a head pose rotation sample from the FacePix database~\cite{little2005methodology}.\label{fig:facepix_3}}
\end{figure}

Specifically, the soft gate mechanism can be formulated as a sigmoid function as follows:
\begin{equation}
\operatorname{S}\left(y\right) =\frac{1}{1+e^{-k\left(\frac{\left\lvert y\right\rvert }{45} -1\right) } } ,\label{eq:10}
\end{equation}
\vspace{2pt}
where $y$ denotes the yaw degree of an input face image within the range of [-90,90], and $k$ is a slope parameter set to 10. In Eq.~\eqref{eq:10}, we take the absolute value of yaw angle to quantify the deviation between frontal-profile poses without considering the left-right profile direction. The yaw coefficient is designed to gradually increases from 0 to 1 while the head pose rotates from frontal to left-right profile as shown in~\Cref{fig:facepix_3}. This enabled the soft gate to control the magnitude of the residual to learn additional features between frontal-profile pose. Note that, although the soft gate is adaptively adjusted according to different face images, the yaw coefficient is a non-trainable parameter during the training process.

\subsection{Channel Attention Module}
In this study, we applied a Channel Attention Module (CAM) to enhance the robustness of DRM. Specifically, we used CAM to improve the representation of interests by reweighting the feature channels after DRM aggregate pose information.

The existing representative works on channel attention mainly difference in squeezing operation that aggregates the global information of a single feature map into a channel descriptor.  For instance, SENet~\cite{hu2018squeeze} generate channel-wise statistics using average-pooling, while CBAM~\cite{woo2018cbam} use both average-pooling and max-pooling signal to learn inter-channel dependencies. We verified the performance of different channel attention methods for PAM. Through extensive ablation studies, we finally adopted the block design of CBAM to aggregate both average-pooling and max-pooling signals for channel attention in PAM (see~\Cref{sec:ablation-study} for more details). In contrast to the aforementioned channel attention approaches, we deliberately placed the CAM after DRM aggregate the pose information to obtain better performance and compatibility with the proposed PAM.

\begin{figure}
  \centering
  \captionsetup[sub]{font=scriptsize}
  \hspace{0.10\linewidth}
  \begin{subfigure}{0.38\linewidth}
    \centering
    \includegraphics[width=0.85\linewidth]{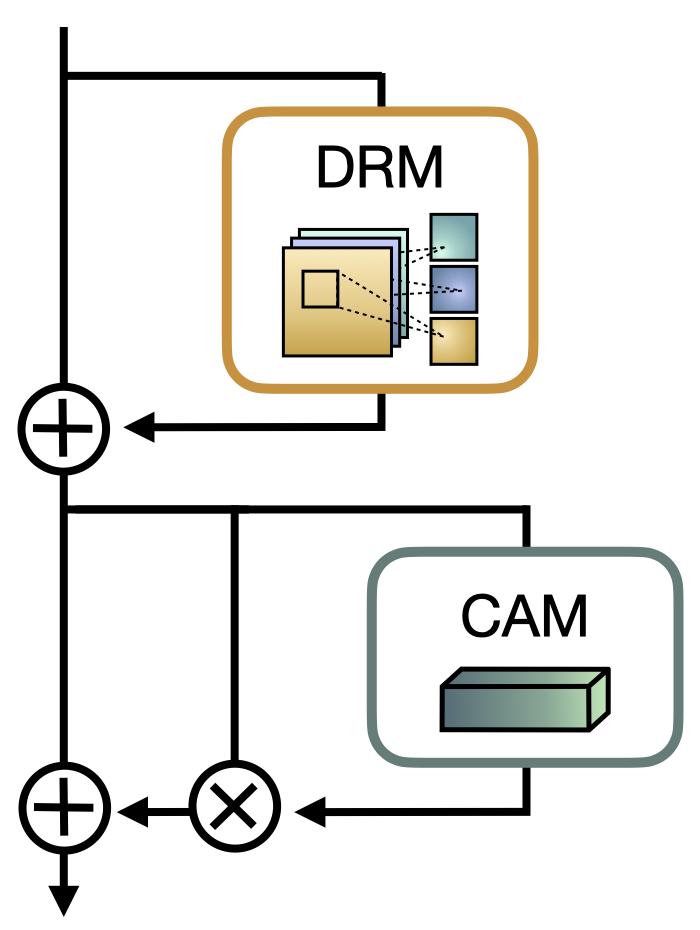}
    \caption{With identity mapping}
    \label{fig:with-identity-mapping}
  \end{subfigure}
  \hfill
  \begin{subfigure}{0.38\linewidth}
    \centering
    \includegraphics[width=0.85\linewidth]{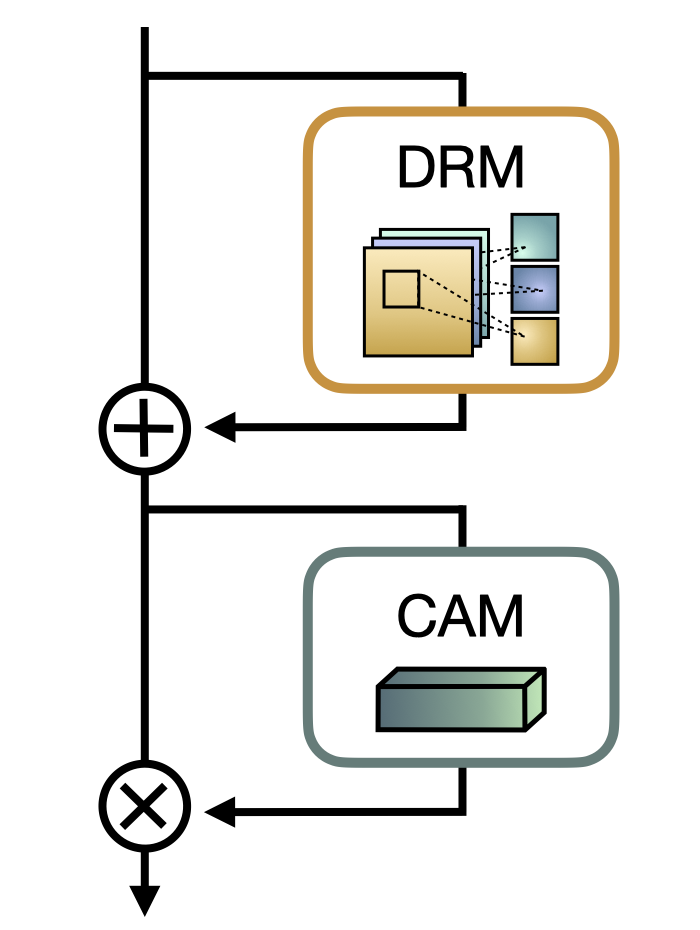}
    \caption{Without identity mapping}
    \label{fig:without-identity-mapping}
  \end{subfigure}
  \hspace{0.10\linewidth}
  \caption{Illustration of applying CAM with or without identity mapping in PAM.}
  \label{fig:with-without-identity-mapping}
\end{figure}

In addition, we also verified the performance of CAM with or without identity mapping, as shown in~\Cref{fig:with-without-identity-mapping}. With identity mapping, CAM performs feature refinement as residual learning. Without identity mapping, CAM directly recalibrates the channel weight of the output feature map from DRM. We observed that performing channel attention without identity mapping obtain better improvement on pose-invariant representation learning (see~\Cref{sec:ablation-study} for more details). Therefore, we used CAM without identity mapping for the final block design of the PAM, as illustrated in~\Cref{fig:without-identity-mapping}.

In summary, our method introduces a Channel Attention Module (CAM) modified from CBAM~\cite{woo2018cbam} that adaptively recalibrates channel-wise feature responses by aggregating both average-pooling and max-pooling signals after the DRM has learned and aggregated the pose information. The final structure of the CAM adopts a `without identity mapping' setting, as shown in~\Cref{fig:channel-attention-module}. Comprehensive experiments demonstrate that CAM can effectively reinforce the representation of interests and achieve better performance for PAM.

\section{Experiments}
\subsection{Implementation Details}
\noindent\textbf{Datasets.} We employ refined MS1MV2~\cite{deng2019arcface} as our training set to conduct fair comparisons with other methods. MS1MV2 contains about 5.8M images of 85K individuals. We extensively test our method on several popular benchmarks, including LFW~\cite{huang2008labeled}, CFP-FP~\cite{sengupta2016frontal}, AgeDB-30~\cite{moschoglou2017agedb}, CPLFW~\cite{zheng2018cross}, and CALFW~\cite{zheng2017cross}.

\noindent\textbf{Experimental Settings.} For the data prepossessing, we follow~\cite{wen2016discriminative,liu2017sphereface,wang2018cosface,deng2019arcface} to generate normalized face cropped (112 × 112) with five facial landmarks. Each pixel (in [0,255]) in RGB images is normalized by subtracting 127.5 and divided by 128. For all the training faces, we only applied random horizontal flip at probability p=0.5 for data augmentation. For the embedding network, we adopt ResNet50 as in~\cite{deng2019arcface}. All experiments are implemented in Pytorch~\cite{NEURIPS2019_9015}. We train models on a single NVIDIA RTX 3090 GPU with a batch size of 256. We employ ArcFace~\cite{deng2019arcface} loss with scaling parameter $s$ set to $64$ and margin $m$ set to $0.5$. The models are trained with SGD algorithm, with momentum $0.9$ and weight decay $5e-4$. On MS1MV2~\cite{deng2019arcface}, the initial learning rate starts from 0.1 and then decreased by a factor of 10 at the 10, 18, 22 epochs. The training process finished at 24 epochs.

\subsection{Ablation Study}
\label{sec:ablation-study}
\noindent\textbf{Examining the Location for PAM.} This experiment compares the effectiveness of inserting PAM at different bottlenecks to investigate the most efficient locations for performing pose-invariant feature transformation hierarchically. In order to facilitate understanding of the PAM placement settings, we add stage numbers of the embedded network after PAM to indicate the placement locations, \textit{e.g.}, the setting of PAM1234 denotes placing PAM at the end of stages 1, 2, 3, 4, respectively. We first explore three different placement strategies: PAM1234, PAM34, and PAM12, which compare the effectiveness of performing pose-invariant feature transformation at all levels, high levels, and low levels of abstraction, respectively. As shown in~\Cref{table:ablation-study-location}, although all placement strategies of PAM outperform the baseline on all benchmarks, we observe that PAM1234 does not gain further improvement due to the additional module insertion compared to PAM12 and PAM34 using only two module settings. Through the comparison between PAM12 and PAM34 placement settings, we also found that PAM12 obtains more performance-boosting than PAM34 on CFP-FP and CPLFW, which indicates that performing pose-invariant feature transformations on low-level features is more effective than on high-level features. This observation is intuitive since feature discrepancy between frontal-profile representation tends to be more correlated to low-level features (e.g., texture and edges) than to the higher-level features used for identification.  Moreover, PAM aims to learn an approximate feature transformation between pose variation that shared by all human faces, which also demonstrate the same conclusion that low-level features are more similar to those features that shared by all human faces (e.g., edges variation due to pose variation, texture variation due to different illumination angle that causes by pose variation).

In addition, we also provide further analysis on the influence of learning higher-level feature transformation to verify the above statement. As shown in~\Cref{table:ablation-study-location}, we train models on PAM123 and PAM124 placement settings. Compared to the PAM12 setting, these settings consist of an additional PAM learning higher-level feature transformations at stage 3 and 4 bottlenecks, respectively. We found that both settings suffer an obvious degradation on CPLFW when adding feature transformation for higher-level features. Even though PAM124 is slightly improved over PAM12 on CFP-FP, the significant degradation on LFW still suggests that higher-level features are not robust for learning pose-invariant feature transformation.

\begin{table}[t]
  \centering
  \renewcommand\arraystretch{1.2}
  \caption{Face verification results (\%) with different PAM placement strategies. We place stage number after PAM to denote different placement strategies, \textit{e.g.}, PAM1234 denote placing PAM at the end of stages 1, 2, 3, and 4, respectively.}
  \resizebox{\linewidth}{!}{
  \begin{tabular}{l|c|c|c|c|c|l}
      \hline
          Methods & {LFW} & {CFP-FP} & {AgeDB-30} & {CPLFW} & {CALFW} & {Params}\\ 
      \hline\hline  
          \text{IR50}           & 99.77 & 97.54 & 97.90 & 92.25 & 95.93 & 43.57M \\ 
      \hline
          \text{IR50 + PAM1234} & 99.80 & 97.54 & 97.90 & 92.67 & 95.97 & {+ 65,600} \\ 
          \text{IR50 + PAM34}   & 99.78 & 97.60 & 98.07 & 92.28 & 95.95 & {+ 58,624} \\ 
          \text{IR50 + PAM12}   & \textbf{99.82} & 97.89 & 98.00 & \textbf{92.80} & 96.05 & \textbf{+ 6,976} \\ 
      \hline
          \text{IR50 + PAM123}  & 99.78 & 97.86 & 98.05 & 92.47 & \textbf{96.10} & {+ 21,056} \\ 
          \text{IR50 + PAM124}  & 99.75 & \textbf{97.91} & \textbf{98.13} & 92.58 & 95.98 & {+ 51,520} \\ 
  \hline
  \end{tabular}}
  \label{table:ablation-study-location} 
\end{table}

To verify the efficiency of the module, we also compared the number of additional parameters for all PAM placement strategies. As shown in~\Cref{table:ablation-study-location}, PAM12 outperforms not only other placement strategies in terms of performance improvement but also superior in negligible parameter and computational overheads. The results demonstrate that the improvement attained by the PAM is not due to the capacity increment but because of the effective block design and placement strategies. Based on the results of this study, we adopt the PAM12 placement strategies in all of the following experiments.

\begin{table}[t]
  \centering
  \renewcommand\arraystretch{1.2}
  \caption{Face verification results (\%) with different convolutional strategies. PAM12-C and PAM12-D refer to using conventional convolution and depthwise convolution in DRM, respectively.} 
  \resizebox{\linewidth}{!}{
\begin{tabular}{l|c|c|c|c|c|l}
      \hline
          Methods & {LFW} & {CFP-FP} & {AgeDB-30} & {CPLFW} & {CALFW} & {Params}\\ 
      \hline\hline  
          \text{IR50}           & 99.77 & 97.54 & 97.90 & 92.25 & 95.93 & 43.57M \\ 
      \hline
          \text{IR50 + PAM12-C} & 99.78 & 97.71 & \textbf{98.02} & \textbf{92.82} & 96.05 & {+ 372,160} \\ 
          \text{IR50 + PAM12-D} & \textbf{99.82} & \textbf{97.89} & 98.00 & 92.80 & \textbf{96.05} & \textbf{+ 6,976} \\
  \hline
  \end{tabular}}
  \label{table:ablation-study-conv} 
\end{table}

\noindent\textbf{Effectiveness of the Depthwise Convolution.}  In this experiment, we investigate the effectiveness of applying depthwise convolution in DRM compared to conventional convolution. As shown in~\Cref{table:ablation-study-conv}, we adopt the PAM12 placement strategy and replace the convolutional layer of DRM with different convolutional strategies. PAM12-C refers to using conventional convolution in DRM, while PAM12-D denotes using depthwise convolution in DRM. The verification results on LFW and CFP-FP demonstrate that the depthwise convolution strategy is superior in terms of performance boost. Although the results on AgeDB-30 and CPLFW are slightly inferior to conventional convolution, the notable differences between the parameters suggest that PAM12-D makes more effective use of the model parameters. Moreover, from the computation process, we can easily find that depthwise convolution is actually a subset of the regular convolution, which zeroes out most of the weights and keeps only one channel working in the convolution kernel. This indicates that depthwise convolution brings more direct and robust learning for pose-invariant feature transformation. Therefore, we adopt the depthwise convolution strategy to build our DRM for module efficiency and better performance enhancement.

\noindent\textbf{Effectiveness of the CAM block design.} In this experiment, we first investigate the squeezing strategies of SENet~\cite{hu2018squeeze} and CBAM~\cite{woo2018cbam} to explore the most compatible strategies for our CAM. Then, we further investigate CAM with different feature refinement settings: with or without identity mapping. The experimental results of the two aspects are summarized in~\Cref{table:ablation-study-cam-design}. We average the results of LFW, CFP-FP, and CPLFW as a composite metric in order to facilitate understanding and comparing performance differences for each experimental setting. As we can see from the results, the CBAM squeezing strategy consistently outperform the SENet squeezing strategy under both feature refinement settings, indicating that the additional max-pooling signal can gather meaningful features to compensate for the average-pooled features. Thus, we use both max-pooling and average-pooling operations for our CAM to achieve better feature refinement for the PAM. On the other hand, the ``without identity mapping" setting is also consistently superior to the ``with identity mapping" settings under both squeezing strategies, demonstrating that directly perform channel weight recalibration on the DRM output feature is more effective in our case. In conclusion, the ``CBAM squeezing strategy" and ``without identity mapping" settings outperform other CAM variants on CPLFW and the average composite metric. Therefore, we eventually adopt these settings as our CAM block design to construct the proposed PAM.

\begin{table}[t]
    \centering
    \renewcommand\arraystretch{1.2}
    \caption{Face verification results (\%) with different CAM block designs.}
    \resizebox{\linewidth}{!}{
	  \begin{tabular}{l|c|c|c|c|c|c}
        \hline
            Methods & {CAM} & \tabincell{c}{Identify\\Mapping} & {LFW} & {CFP-FP} & {CPLFW} & {Average}\\ 
        \hline\hline  
            \multirow{4}{*}{IR50 + PAM12}
            & SE    &  w  & 99.82 & 97.57 & 92.37 & 96.59\\ 
            & SE    & w/o & 99.78 & \textbf{98.01} & 92.48 & 96.76\\ 
            & CBAM  &  w  & 99.82 & 97.57 & 92.72 & 96.70\\ 
            & CBAM  & w/o & \textbf{99.82} & 97.89 & \textbf{92.80} & \textbf{96.84}\\ 
		\hline
    \end{tabular}}
    \label{table:ablation-study-cam-design} 
\end{table}

\noindent\textbf{Effectiveness of the Soft Gate.} As mentioned in~\Cref{sec:softgate}, the soft gate mechanism of PAM controls the magnitude of the residuals by generating different yaw coefficients to learn feature discrepancies between frontal-profile pose variations. If we construct PAM without the soft gate mechanism, DRM will degenerate into another residual unit that composed ResNet variants, which increases the model depth but loses the learning orientation in terms of pose variations. This means that the proposed PAM should learn better pose-invariant representation under the soft gate mechanism; otherwise, the improvement achieved could be the result of adding layers. Therefore, we conduct an ablation study on the soft gate mechanism to verify the aforementioned statement. In this experiment, we set the soft gate consistently to a constant of 1 for all samples, i.e., $S(y) = 1$, to represent the removal of the soft gate mechanism.

As reported in~\Cref{table:ablation-study-yaw}, we observe an obvious performance gap on CFP-FP and CPLFW between the proposed PAM and the ``without soft gate" one. Even though they keep an identical amount of trainable parameters, the proposed PAM still outperforms the ``without soft gate" PAM on most of the challenging benchmarks. This demonstrates that the soft gate mechanism plays an essential role in providing pose-invariant learning orientation for the PAM. The significant improvement of PAM is mainly due to its effective block design. Therefore, we use a soft gate as an attention mechanism to emphasize informative features of pose variations in the proposed PAM to obtain better pose-invariant representations.

\begin{table}[t]
    \centering
    \renewcommand\arraystretch{1.2}
    \caption{Face verification results (\%) with different soft gate strategies. (w/o soft gate refers to replacing the yaw coefficient with an identity mapping function, S(y)=1)}
    \resizebox{\linewidth}{!}{
	\begin{tabular}{l|c|c|c|c|c|c}
        \hline
            Methods & {Soft Gate} & {LFW} & {CFP-FP} & {AgeDB-30} & {CPLFW} & {CALFW} \\ 
        \hline\hline  
            \multirow{2}{*}{IR50 + PAM12} & w/o & 99.80 & 97.64 & \textbf{98.02} & 92.67 & 95.85 \\ 
            & w & \textbf{99.82} & \textbf{97.89} & 98.00 & \textbf{92.80} & \textbf{96.05} \\
		\hline
    \end{tabular}}

    \label{table:ablation-study-yaw} 
\end{table}

\begin{table}[t]
  \centering
  \renewcommand\arraystretch{1.2}
  \caption{Face verification results (\%) with different PAM block designs.}
  \resizebox{\linewidth}{!}{
\begin{tabular}{l|c|c|c|c|c|c|c}
      \hline
      \multirow{2}{*}{Methods} & \multicolumn{2}{c|}{Submodule} & \multirow{2}{*}{LFW} & \multirow{2}{*}{CFP-FP} & \multirow{2}{*}{AgeDB-30} & \multirow{2}{*}{CPLFW} & \multirow{2}{*}{CALFW}\\ \cline{2-3}
      & {DRM} & {CAM} &&&&& \\
      \hline\hline  
          \text{IR50} & & & 99.77 & 97.54 & 97.90 & 92.25 & 95.93\\
      \hline 
          \multirow{3}{*}{IR50 + PAM12}
          & \checkmark & & 99.82 & 97.63 & \textbf{98.02} & 92.42 & \textbf{96.10} \\
          &  & \checkmark & 99.80 & 97.81 & 97.95 & 92.58 & 95.97 \\
          & \checkmark & \checkmark & \textbf{99.82} & \textbf{97.89} & 98.00 & \textbf{92.80} & 96.05 \\ 
  \hline
  \end{tabular}}
  \label{table:ablation-study-with-DRM-CAM} 
\end{table}

\noindent\textbf{Effectiveness of the DRM and CAM.} In this experiment, we explore the essentiality of each submodule in the proposed PAM. With the interaction of DRM and CAM, the proposed PAM can enhance the learning process to obtain better pose-invariant representations. Therefore, we conduct an ablation study on the final block design of PAM to verify the effectiveness of the DRM and CAM, respectively. ~\Cref{table:ablation-study-with-DRM-CAM} summarizes the experimental results. We compare different submodule settings, including DRM-only, CAM-only, and DRM+CAM settings. As we can see from the results, all settings show consistent improvements on all benchmarks compared to the IR50 baseline. The performance improvement on the DRM-only settings demonstrates the effectiveness of DRM in learning feature discrepancy between pose variations. Through the comparison between DRM-only and DRM+CAM settings, we observe obvious performance improvements on LFW, CFP-FP, and CPLFW when CAM is involved, indicating that CAM plays an essential role in improving the representation of interests after DRM aggregate pose information. Moreover, we also compare DRM+CAM settings with the CAM-only results. As we can see from the results, the performance of PAM degrades as we remove DRM from PAM. This suggests that the improvement attained by PAM is not merely due to the benefits of channel attention. DRM has a fundamental role to play in learning pose-invariant features in PAM. In conclusion, with the interaction of DRM and CAM, the performance of the proposed PAM can be optimized in terms of learning better pose-invariant representations.

\subsection{Evaluation Results}
\noindent\textbf{Comparison with Arcface.} Recently, ArcFace~\cite{deng2019arcface} has achieved state-of-the-art performance on several challenging face recognition benchmarks. As shown in~\Cref{table:comapare-result-to-arcface}, they report the performance compared to other open-sourced face recognition models on CPLFW and CALFW datasets to verify the robustness under extremely challenging pose and age variations. The results of the ArcFace model are trained on a 100-layer ResNet variant using the MS1MV2 training set. Due to hardware limitations, we report our performance on the IR50 embedding network, which is identical to the ResNet variant of ArcFace but consists of only 50 layers. As we can see from the results, our approach achieves remarkable performance enhancement on the challenging pose variations and age variations benchmarks compared to ArcFace. In addition, the proposed PAM outperforms state-of-the-art face recognition models by using only about half the number of layers of Arcface, which indicates that the PAM block design is extremely effective under challenging pose variations.

\begin{table}[t]
    \centering
    \renewcommand\arraystretch{1.2}
    \caption{Face verification results (\%) of open-sourced face recognition models on LFW, CALFW and CPLFW.}
    \resizebox{\linewidth}{!}{
    \begin{tabular}{l|c|c|c|c}
        \hline 
            Methods & LFW & CALFW & CPLFW & Params\\
        \hline\hline
            HUMAN-Individual & 97.27 & 82.32 & 81.21 & -\\
            HUMAN-Fusion     & 99.85 & 86.50 & 85.24 & -\\
        \hline
            Center Loss~\cite{wen2016discriminative}       & 98.75 & 85.48 & 77.48 & -\\
            SphereFace~\cite{liu2017sphereface}            & 99.27 & 90.30 & 81.40 & -\\
            VGGFace2~\cite{cao2018vggface2}                & 99.43 & 90.57 & 84.00 & -\\
            ArcFace (MS1MV2, IR100)~\cite{deng2019arcface} & 99.82 & 95.45 & 92.08 & 65.13M \\
        \hline
            Ours (MS1MV2, IR50 + PAM12) & \textbf{99.82} & \textbf{96.05} & \textbf{92.80} & \textbf{43.58M}\\
        \hline
    \end{tabular}}
    \label{table:comapare-result-to-arcface} 
\end{table}

\noindent\textbf{Comparison with DREAM Block.} We compare our method with the representative pose-invariant methods, DREAM~\cite{cao2018pose}, which has the same lightweight modularity advantage of solving pose-invariant feature transformation in the deep feature space. We reproduce the IR50 architectures as the baseline embedding network for both the DREAM block and the proposed PAM in order to conduct fair comparison. As shown in~\Cref{table:comapare-result-to-DREAM}, both PAM and DREAM block significantly improve all evaluation benchmarks by an obvious margin compared to the IR50 baseline. The results of CFP-FP and CPLFW evaluations demonstrate that PAM outperforms the DREAM block on the performance of performing pose-invariant feature transformation. Furthermore, we also observe a significant improvement in reducing module parameters by comparing the number of parameters between DREAM and PAM. The proposed PAM effectively reduces memory requirements by more than a factor of 75 compared to DREAM. This indicates that the deliberate block design of PAM not only outperforms the DREAM block in pose feature transformation but also extremely memory efficient that can implemented with a negligible burden. Although the results on AgeDB-30 and CALFW are slightly inferior to DREAM, the noticeable difference between the parameters and the superior results on CFP-FP and CPLFW demonstrate that PAM makes more effective use of model parameters focused on learning pose-invariant feature transformation rather than learning other unknown features.

\begin{table}[t]
    \centering
    \renewcommand\arraystretch{1.25}
	\caption{Face verification results (\%) of lightweight pose-invariant methods on LFW, CFP-FP, AgeDB-30, CALFW and CPLFW.}
  \resizebox{\linewidth}{!}{
    \begin{tabular}{l|c|c|c|c|c|l}
        \hline
            Methods & {LFW} & {CFP-FP} & {AgeDB-30} & {CPLFW} & {CALFW} & {Params}\\ 
        \hline\hline  
            \text{IR50}           & 99.77 & 97.54 & 97.90 & 92.25 & 95.93 & 43.57M \\ 
        \hline
            \text{IR50 + DREAM}   & 99.82 & 97.60 & \textbf{98.05} & 92.65 & \textbf{96.08} & {+ 525,312} \\ 
            \text{IR50 + PAM12}   & \textbf{99.82} & \textbf{97.89} & 98.00 & \textbf{92.80} & 96.05 & \textbf{+ 6,976} \\ 
		\hline
    \end{tabular}}
    
    \label{table:comapare-result-to-DREAM} 
\end{table}

\section{Conclusions}
We proposed a Pose Attention Module (PAM) to enhance pose-robust feature learning for deep face recognition. PAM is a lightweight and easy-to-implement attention block that performs frontal-profile feature transformation in hierarchical feature space by learning the residuals between pose variations with a soft gate mechanism. We observed that feature transformation in hierarchical feature space is more effective than in feature embedding space, as it can take advantage of different feature levels to better learn feature transformation and benefit from joint learning with representation learning. Experimental results show that our method not only outperforms state-of-the-art methods but also effectively reduces memory requirements by more than 75 times. It is noteworthy that our method is not limited to face recognition with large pose variations. By adjusting the soft gate mechanism of PAM to a specific coefficient, such attention block can easily extend to address other intra-class imbalance problems in face recognition, including large variations in age, illumination, expression, etc.


{\small
\bibliographystyle{ieee_fullname}
\bibliography{egbib}
}

\end{document}